\documentclass{article} 
\usepackage{iclr2027_conference,times}

\usepackage{amsmath,amsfonts,bm}

\def\eqref#1{equation~\ref{#1}}

\def\1{\bm{1}}

\DeclareMathAlphabet{\mathsfit}{\encodingdefault}{\sfdefault}{m}{sl}
\SetMathAlphabet{\mathsfit}{bold}{\encodingdefault}{\sfdefault}{bx}{n}

\usepackage{hyperref}
\usepackage{url}
\usepackage{graphicx}
\usepackage{subcaption}
\usepackage[normalem]{ulem}
\usepackage{booktabs}
\usepackage{multirow}
\usepackage{algorithm}
\usepackage{algpseudocode}

\DeclareRobustCommand{\revdel}[1]{%
    \bgroup
    \markoverwith{\textcolor{red}{\rule[0.5ex]{2pt}{0.4pt}}}%
    \ULon{#1}%
}

\newcommand{\method}{\textsc{ObsSpec}}

\title{\method: Accelerating Agentic Reinforcement Learning with Observation Speculation}

\author{%
\makebox[\dimexpr\textwidth-2\tabcolsep\relax][c]{\begin{tabular}[t]{@{}c@{}}
Kyle Montgomery$^{1,2}$\thanks{Work done during an internship at ByteDance Seed.}
, Wang Zhang$^{2}$, Chenguang Wang$^{1}$\thanks{Corresponding author.} \\
\normalfont $^{1}$UC Santa Cruz, $^{2}$ByteDance Inc. \\
\normalfont \texttt{\{kylemontgomery,chenguangwang\}@ucsc.edu}
\end{tabular}}%
}

\iclrfinalcopy 
\begin{document}

\maketitle
\lhead{}
\renewcommand{\headrulewidth}{0pt}

\begin{abstract}
Tool-using agents alternate between model generation and tool execution, waiting for each tool call to complete before starting the next generation. This sequential execution is costly during agentic reinforcement learning (RL), which requires large batches of multi-turn rollouts. We introduce \method, an observation speculation pipeline for agentic RL that uses a small world model to predict tool outputs while the environment executes, allowing the policy to begin its next generation early. \method\ retains speculative work only when the predicted and actual observations match exactly, making rollout collection lossless, while co-training keeps the world model up to date as the policy learns. On terminal-use tasks, predictions from a 0.6B world model exactly match actual observations in 66.4\% of attempts and arrive roughly 16 times faster. Using these predictions, \method\ reduces agent latency by 9.3\% and 14.2\% for 8B and 32B policies. Across controlled experiments, we characterize the regimes in which \method\ reduces agent latency under shared policy resources. Finally, because \method\ overlaps policy generation with tool execution, it is architecturally compatible with existing techniques for accelerating RL. Code is available at \url{https://github.com/kylemontgomery1/verl-recipe/tree/obsspec/obsspec}.
\end{abstract}

\section{Introduction}
Tool-using agents solve complex tasks by alternating between model generation and tool execution~\citep{yao2022react,schick2023toolformer,qin2024toolllm}. A terminal agent, for example, generates a shell command, waits for a sandbox to execute it, observes the output, and then generates its next command. In our experiments with terminal agents, tool execution makes up nearly half of the agent's latency. The cost is significant during agentic reinforcement learning (RL), where each policy update requires hundreds or even thousands of concurrent, multi-turn rollouts.

Observation speculation predicts a tool's output while the tool executes, allowing the policy to begin generating its next action before the actual observation arrives. Recent works apply this technique at inference time, substantially reducing agent latency~\citep{ye2025speculative,cao2026auton,saberi2026spechop,li2026speculate,chen2026aospec}.

Applying observation speculation to RL introduces three requirements. First, rollout collection must be lossless: rollouts contain no mispredicted observations or actions generated from them. Second, speculation should accelerate rollout collection under shared policy resources. Because many rollouts run concurrently, speculative policy generations can slow other rollouts by competing for the same policy resources. Third, predictions must remain accurate as the policy learns. Policy updates can change the rollout distribution, so a frozen world model may become less accurate as training progresses.

\begin{figure*}[t]
    \centering
    \includegraphics[width=0.358\textwidth]{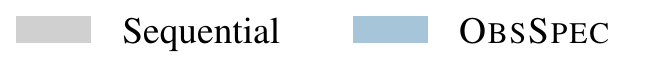}\\[-0.25em]
    \captionsetup{skip=4pt}
    \captionsetup[subfigure]{justification=centering,skip=2pt}
    \begin{minipage}[t]{0.24\textwidth}
        \centering
        \includegraphics[width=\linewidth]{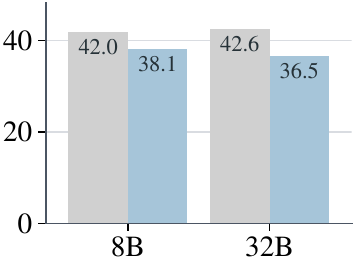}
    \end{minipage}
    \hfill
    \begin{minipage}[t]{0.24\textwidth}
        \centering
        \includegraphics[width=\linewidth]{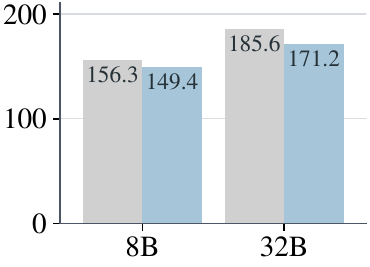}
    \end{minipage}
    \hfill
    \begin{minipage}[t]{0.24\textwidth}
        \centering
        \includegraphics[width=\linewidth]{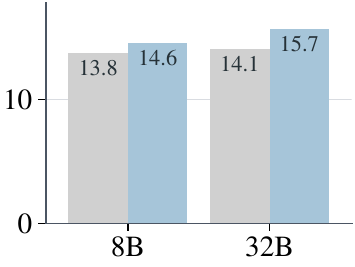}
    \end{minipage}
    \hfill
    \begin{minipage}[t]{0.24\textwidth}
        \centering
        \includegraphics[width=\linewidth]{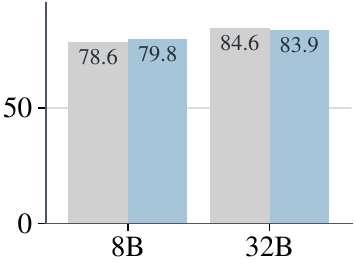}
    \end{minipage}
    \par\vspace{-0.10em}
    {\small Policy size}
    \par\vspace{-0.65em}
    \subcaptionbox{Agent latency (s) $\downarrow$.}[0.24\textwidth]{\rule{0pt}{0pt}}
    \hfill
    \subcaptionbox{Batch latency (s) $\downarrow$.}[0.24\textwidth]{\rule{0pt}{0pt}}
    \hfill
    \subcaptionbox{Async throughput\\(k rollouts/h) $\uparrow$.}[0.24\textwidth]{\rule{0pt}{0pt}}
    \hfill
    \subcaptionbox{Reward (\%) $\uparrow$.}[0.24\textwidth]{\rule{0pt}{0pt}}
    \caption{Across policy sizes, \method\ reduces latency and increases asynchronous rollout throughput with similar rewards.}
    \label{fig:main-results}
\end{figure*}

We introduce \method, an observation speculation pipeline for agentic RL designed around these requirements. A small world model predicts the next observation while the environment executes the corresponding action, allowing the policy to begin its next generation as soon as the prediction completes, without waiting for tool execution to finish. When the actual observation arrives, we retain the speculative generation only if the predicted and actual observations match exactly and discard it otherwise. This keeps rollout collection lossless. To limit contention with non-speculative policy generations, we bound how far ahead the pipeline speculates (depth) and how many alternative observations it predicts (breadth). We evaluate how these limits affect rollout latency under shared policy resources. Finally, we co-train the world model with the policy, using actual observations as supervision to keep its predictions accurate as training proceeds.

Experiments on terminal-use tasks show that \method\ predicts tool outputs quickly and accurately. For example, predictions from a 0.6B world model exactly match actual observations in 66.4\% of attempts and arrive roughly 16 times faster than tool execution. \method\ reduces agent latency by 9.3\% and 14.2\% for 8B and 32B policies, respectively, and increases asynchronous rollout throughput by up to 11.5\%. By overlapping policy generation with tool execution, \method\ is architecturally compatible with other RL acceleration techniques, including fully asynchronous rollout scheduling~\citep{fu2026areal,piche2025pipelinerl}, sample-efficient RL algorithms~\citep{chen2025minimax,gao2025soft, shrivastava2026echo}, low-precision inference~\citep{xiao2023smoothquant,qiu2026fp8}, speculative decoding~\citep{leviathan2023fast,chen2026respec,pham2026specroll}, and multi-token prediction~\citep{gloeckle2024better,wang2026mtp}.

In short, we make three contributions.
\begingroup
\setlength{\leftmargini}{1.5em}
\begin{itemize}
    \item We introduce \method, a lossless observation speculation pipeline for agentic RL that co-trains a lightweight world model with the policy.
    \item We show that tool outputs can often be predicted accurately and substantially faster than the corresponding tool calls can be executed. \method\ translates this opportunity into up to 14.2\% lower agent latency and 11.5\% higher asynchronous rollout throughput on terminal-use tasks.
    \item We show how world model size, speculation depth, and speculation breadth affect rollout latency under shared policy resources, and how co-training helps maintain prediction accuracy as the policy learns.
\end{itemize}
\endgroup

\section{Approach}
\label{sec:approach}

\subsection{Problem Setup}
\label{sec:rollout-setting}

To solve a task $x$, a tool-using agent takes a sequence of actions toward the task's goal. Each action is a tool call, and its output is an observation. At time step $t$, the agent conditions on the interaction history $h_t=(x,a_1,o_1,\ldots,a_{t-1},o_{t-1})$ and samples an action $a_t$. The environment $\mathcal{E}$ then executes $a_t$, updates its state from $s_t$ to $s_{t+1}$, and returns an observation $o_t$:
\begin{equation}
    a_t\sim\pi_\theta(\cdot\mid h_t), \qquad (s_{t+1},o_t)\leftarrow\mathcal{E}(s_t,a_t).
\end{equation}

For instance, in a terminal environment, $a_t$ is a tool call containing a shell command, $s_t$ includes the current filesystem and process state, and $o_t$ contains the stdout and stderr returned by the sandbox. A complete sequence $\tau=(x,a_1,o_1,\ldots,a_T,o_T)$ constitutes a rollout and receives a reward $R(\tau)$ from a verifier. During RL, the current policy $\pi_{\theta_k}$ collects a batch of rollouts and uses their rewards in a policy update.

\begin{figure*}[t]
    \centering
    \includegraphics[width=\textwidth,trim=31 256 95 327,clip]{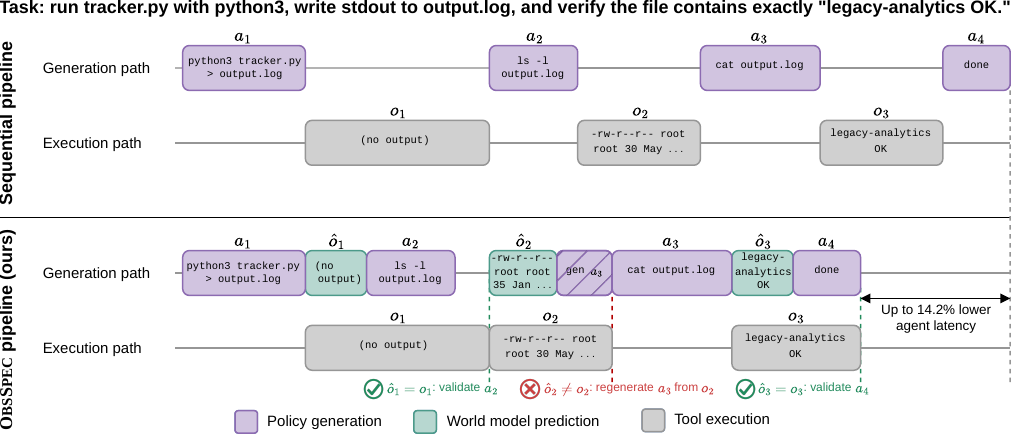}
    \caption{Comparison of the sequential and \method\ pipelines on a terminal task. In the sequential pipeline (top), policy generation and tool execution alternate, so generation of each action cannot begin until the previous action finishes. In our \method\ pipeline (bottom), a world model predicts each observation while the corresponding action executes, allowing the policy to generate the next action before the execution of the previous action finishes. The first predicted observation $\hat{o}_1$ matches the actual observation $o_1$, whereas $\hat{o}_2$ differs from $o_2$. Consequently, the in-progress generation of $a_3$ is canceled and restarted from the actual observation $o_2$. By overlapping policy generation with tool execution, \method\ reduces agent latency by up to 14.2\%.}
    \label{fig:speculative-pipeline}
\end{figure*}

In the standard sequential pipeline shown in Figure~\ref{fig:speculative-pipeline}, the policy cannot generate $a_{t+1}$ until the tool has executed $a_t$ and returned $o_t$, so policy generation and tool execution alternate on the critical path. Often, the time spent waiting on tool execution is significant. For example, on terminal-use tasks from Endless Terminals~\citep{gandhi2026endless}, nearly half of agent latency (47.1\%) is spent waiting for sandbox execution.

\subsection{Speculating Tool Outputs}
\label{sec:observation-speculation}
Tool outputs are often highly predictable. In a terminal setting, for example, many commands such as changing directories return no output, while others repeat information already available in the history, such as the contents of a previously viewed file. If an observation can be predicted before the corresponding tool call finishes executing, the policy could begin generating its next action without waiting for the tool output. Recent works exploit this opportunity during inference through observation speculation: they predict tool outputs and start the next policy generation before the tool call finishes executing~\citep{ye2025speculative,cao2026auton,saberi2026spechop,li2026speculate,chen2026aospec}. We study how to adapt this inference-time technique to an RL setting.

Applying observation speculation to RL introduces three requirements. First, lossless rollout collection requires that a rollout contain no mispredicted observations or actions generated from them. Existing inference-time methods for observation speculation may accept semantically similar observations~\citep{saberi2026spechop} or reuse speculative tool execution when the next action matches~\citep{ye2025speculative,li2026speculate}, which can violate this requirement and corrupt training. Second, speculation should accelerate rollout collection under shared policy resources. Existing methods assume request latencies are independent, and evaluate proprietary models such as GPT-5, where large deployments absorb speculative generations without noticeably slowing other requests~\citep{ye2025speculative,li2026speculate}, or estimate latency savings through simulation or trajectory replay~\citep{saberi2026spechop,chen2026aospec}. In contrast, RL systems generate many rollouts concurrently to make full use of the allocated policy resources~\citep{sheng2025hybridflow,fu2026areal,piche2025pipelinerl}. Consequently, speculative requests compete with non-speculative requests for the same resources, so the gains from speculation must offset the slowdown in policy generation. Third, predictions must remain accurate as the policy learns. Policy updates can change the tool calls the agent makes, the states it visits, and the observations it receives. A world model fit to an earlier distribution may consequently become less accurate, making speculation less effective.

To address these challenges, we propose \method, an observation speculation pipeline shown in Figure~\ref{fig:speculative-pipeline}. After the policy samples $a_t$, the pipeline starts executing it in the environment and simultaneously asks a small world model $q_\phi$ to predict the resulting observation $\widehat{o}_t \sim q_\phi(\cdot\mid h_t,a_t)$. If tool execution finishes first, we cancel the prediction, append $o_t$ to the history, and sample $a_{t+1}\sim\pi_\theta(\cdot\mid h_{t+1})$, where $h_{t+1}=(h_t,a_t,o_t)$. If the prediction finishes first, we form the speculative history $\widehat{h}_{t+1}=(h_t,a_t,\widehat{o}_t)$ and begin sampling $a_{t+1}\sim\pi_\theta(\cdot\mid\widehat{h}_{t+1})$ while the tool continues executing.

When the actual observation arrives, we compare it with the prediction. If $\widehat{o}_t=o_t$, we retain the speculative generation of $a_{t+1}$, waiting for it to finish if necessary. Otherwise, we discard that generation and regenerate $a_{t+1}$ from the actual history $h_{t+1}$. Because actions can mutate the environment state, we execute them sequentially, checking each prediction before executing the next action. This procedure ensures that rollout collection is lossless: the rollout contains no mispredicted observations or actions generated from them. A wrong prediction therefore wastes speculative computation without corrupting the rollout. Pseudocode for \method\ is in Algorithm~\ref{alg:wmspec}.

We use speculation depth and breadth to limit how much speculative work the pipeline generates. These limits matter because concurrent RL rollouts share a finite amount of policy resources~\citep{sheng2025hybridflow,fu2026areal,piche2025pipelinerl}. Speculation depth $D$ allows speculative work to advance by alternating predicted observations and conditional policy actions, with at most $D$ actions generated ahead of actual execution~\citep{saberi2026spechop}. Speculation breadth $K$ instead produces up to $K$ candidate predicted observations and launches a speculative policy generation from each distinct candidate~\citep{ye2025speculative,li2026speculate,chen2026aospec}. Increasing depth or breadth creates more opportunities for overlap, but also requires more policy generations, which can slow other rollouts even when predictions are correct. We evaluate this tradeoff in Section~\ref{sec:configuration-results}.

\subsection{Co-Training the Policy and World Model}
\label{sec:world-model-training}
To keep predictions accurate as the policy learns, we co-train the policy and world model on the same rollouts. Each rollout provides rewards for the policy update (e.g., via GRPO~\citep{shao2024deepseekmath, yu2026dapo}) and actual observations for world model training. Given a rollout tokenized as $z_{1:N}$, we train the world model with the masked cross-entropy loss
\begin{equation}
    \mathcal{L}_{\mathrm{WM}}(\phi)
    =
    -\frac{1}{\sum_i m_i}
    \sum_{i=1}^{N}
    m_i\log q_\phi(z_i\mid z_{<i}).
\end{equation}
Here, $m_i$ is a mask indicating whether $z_i$ is an observation token. We include the end-of-observation token (e.g., \texttt{</tool\_response>}) in the mask so that it serves as the stop token during world model generation.

At training iteration $k$, we collect a batch of rollouts using policy $\pi_{\theta_k}$ and world model $q_{\phi_k}$. Using this batch, the policy and world model are updated concurrently: rollout rewards update the policy to $\pi_{\theta_{k+1}}$, while the actual observations update the world model to $q_{\phi_{k+1}}$ using $\mathcal{L}_{\mathrm{WM}}$. Co-training lets the world model track the policy's changing rollout distribution.

\section{Experiments}

\subsection{Experimental Setup}
\label{sec:experimental-setup}

\paragraph{Agent.}
Each task provides the agent with a natural-language instruction and two tools: \texttt{bash} and \texttt{done}. The \texttt{bash} tool executes a shell command in a stateful sandbox and returns its stdout, stderr, and exit code. When the agent believes the task is complete, it calls \texttt{done}.

\paragraph{Tasks.}
Our main experiments evaluate \method\ on long-horizon terminal-use tasks from Endless Terminals~\citep{gandhi2026endless}. For example, the task in Figure~\ref{fig:speculative-pipeline} asks the agent to execute a Python program, redirect its output to a file, and verify the file's contents. Solving such tasks may require inspecting existing files, executing programs, transforming their outputs, and checking the resulting state across multiple shell commands. Each task includes test cases that inspect the final sandbox state and award a binary reward.

\paragraph{Policy and World Models.}
We use Qwen3-8B and Qwen3-32B as policies and Qwen3-0.6B as the world model~\citep{yang2025qwen3}. Unless otherwise noted, policies are trained for 500 steps with GRPO~\citep{shao2024deepseekmath, yu2026dapo}. Specifically, we train each policy and save its rollouts, and then replay the rollouts in the same order to train world models offline. This procedure produces the same artifacts as the co-training procedure outlined in Section~\ref{sec:world-model-training}, but allows us to compare speculation configurations under the same fixed policy, isolating the effect of speculation from differences in policy behavior.

\paragraph{Speculation configuration.}
Unless otherwise specified, \method\ uses depth $D=1$ and breadth $K=1$. The sequential baseline uses the same policy, sampling parameters, evaluation prompts, and policy inference resources, but waits for every tool output before beginning the next policy generation.

\begin{figure*}[t]
    \centering
    \captionsetup{skip=4pt}
    \captionsetup[subfigure]{justification=centering,skip=2pt}
    \begin{subfigure}[t]{\textwidth}
        \centering
        \includegraphics[width=\linewidth]{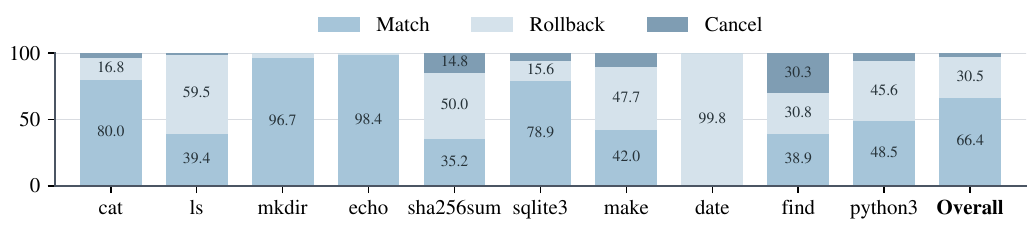}
        \caption{Outcomes (\%) by command.}
    \end{subfigure}
    \vspace{0em}
    \begin{subfigure}[t]{\textwidth}
        \centering
        \includegraphics[width=\linewidth]{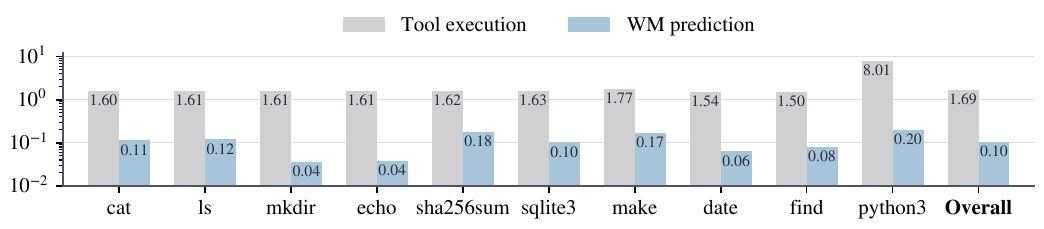}
        \caption{Latency (s) by command (log scale).}
    \end{subfigure}
    \caption{Observation prediction outcomes (top) and latency (bottom) for common commands.}
    \label{fig:command-predictability}
\end{figure*}

\paragraph{Evaluation protocol and metrics.}
Our standard protocol mimics synchronous RL rollout generation with 4,096 rollouts partitioned into 16 batches. Each batch contains 16 held-out prompts and 16 sampled rollouts per prompt; batch assignments are identical across all experiments. The primary metric in these experiments is agent latency: wall-clock time spent generating policy actions and executing tool calls, excluding sandbox setup and final verification. Batch latency and asynchronous rollout throughput are measured on a larger set of 65,536 rollouts, and unlike agent latency, both include sandbox setup and verification time. We also measure the mean reward obtained by the policy. We classify each speculation attempt into three mutually exclusive outcomes: \emph{match}, when the predicted observation exactly equals the actual observation; \emph{rollback}, when the observations differ; and \emph{cancel}, when the tool finishes before the world model. To quantify policy inference contention and wasted computation, we report per-request decode rate and discarded tokens for the policy. For each metric, we report the mean and standard error in Appendix~\ref{sec:full-results}.

\paragraph{Serving and sandbox infrastructure.}
We use \textsc{verl} to orchestrate training and evaluation workloads~\citep{sheng2025hybridflow}. Policies and world models are served asynchronously with vLLM~\citep{kwon2023efficient}. The 8B and 32B policies are allocated eight and sixteen GPUs, respectively, and world models are allocated their own eight GPUs. Most experiments use our default sandbox service. We additionally evaluate Modal and Daytona in Section~\ref{sec:sandbox-results}.

\subsection{Observation Predictions Are Fast and Often Match}
\label{sec:command-predictability}

Observation speculation can help only when tool outputs are both predictable and predicted before tool execution finishes. We first evaluate these two prerequisites using the 8B policy and 0.6B world model. Across 47,762 speculation attempts, 66.4\% match exactly, 30.5\% roll back, and only 3.1\% are canceled because execution finishes first. Thus, the world model finishes before the tool in 96.9\% of attempts and exactly matches the actual observation in approximately two thirds of attempts. Among completed predictions, mean world model latency is 0.10 seconds, compared with 1.69 seconds for tool execution. These results establish that many tool outputs can be predicted accurately and substantially faster than the corresponding tool calls can be executed.

Figure~\ref{fig:command-predictability} demonstrates this across a selection of common shell commands. \texttt{cat}, which accounts for 40.9\% of tool calls, has an 80.0\% match rate because files are often created or inspected earlier in the rollout. Moreover, outputs from commands which normally return no output, such as \texttt{mkdir}, are highly predictable and much faster to predict than to obtain through tool execution. On the other hand, outputs from commands such as \texttt{date} are nearly impossible to predict. However, predictability alone does not determine speculation value. Python commands have a match rate of only 48.5\%, but the gap between execution latency and prediction latency is significant, leaving more time to generate the next action while the tool executes. Conversely, a command can produce a highly predictable output but be too fast or too close to the end of a rollout to materially affect latency.

\subsection{\method{} Improves Rollout Latency and Throughput}
\label{sec:main-results}

We next evaluate whether \method\ converts fast, accurate observation predictions into lower agent latency and higher rollout throughput. Figure~\ref{fig:main-results} summarizes the results using a small 0.6B world model. With the 8B policy, \method\ reduces agent latency by 9.3\% and synchronous batch latency by 4.4\%, and increases asynchronous rollout throughput by 5.9\%. Similarly, with the 32B policy, \method\ reduces agent latency by 14.2\% and synchronous batch latency by 7.8\% and increases asynchronous rollout throughput by 11.5\%. The mean rewards are similar across pipelines, as expected because the rollout collection is lossless. We hypothesize that the larger gains with the 32B policy arise because slower generation would leave more policy work available to overlap with tool execution.

\begin{figure*}[t]
    \centering
    \captionsetup{skip=4pt}
    \captionsetup[subfigure]{justification=centering,skip=2pt}
    \begin{minipage}[t]{0.40\textwidth}
        \centering
        \includegraphics[width=\linewidth]{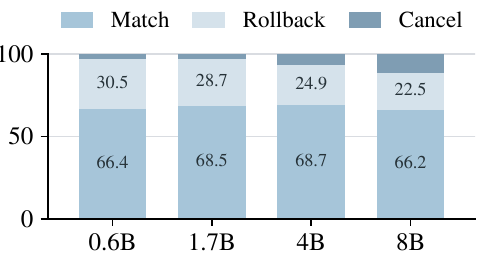}
    \end{minipage}
    \hfill
    \begin{minipage}[t]{0.28\textwidth}
        \centering
        \includegraphics[width=\linewidth]{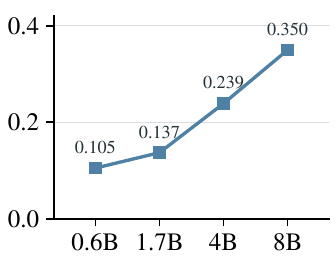}
    \end{minipage}
    \hfill
    \begin{minipage}[t]{0.28\textwidth}
        \centering
        \includegraphics[width=\linewidth]{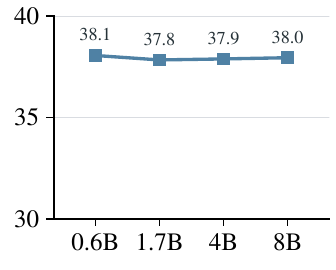}
    \end{minipage}
    \par\vspace{-0.50em}
    {\small World model size}
    \par\vspace{-0.65em}
    \subcaptionbox{Prediction outcomes.}[0.40\textwidth]{\rule{0pt}{0pt}}
    \hfill
    \subcaptionbox{WM latency (s).}[0.28\textwidth]{\rule{0pt}{0pt}}
    \hfill
    \subcaptionbox{Agent latency (s).}[0.28\textwidth]{\rule{0pt}{0pt}}
    \caption{Effect of world model size with the 8B policy at $D=K=1$. Larger models reduce rollbacks but take longer to predict and do not reduce agent latency.}
    \label{fig:wm-size}
\end{figure*}

\subsection{Larger World Models Do Not Reduce Agent Latency}
\label{sec:world-model-size-results}

Larger world models reduce rollbacks but take longer to generate predictions.  Figure~\ref{fig:wm-size} demonstrates this trade-off across four world models ranging from 0.6B to 8B parameters. As we scale the world model, rollbacks fall from 30.5\% to 22.5\%. At the same time, mean prediction latency increases from 0.105 to 0.350 seconds, and the cancel rate rises from 3.1\% to 11.4\% because tool execution is more likely to finish before the prediction. These effects largely cancel, with all four world models producing nearly identical agent latencies. We therefore prefer the 0.6B world model because it provides a similar reduction in agent latency for substantially less compute.

\subsection{Less Is More Under Shared Policy Resources}
\label{sec:configuration-results}

\paragraph{Speculation depth.}

\begin{figure*}[t]
    \centering
    \captionsetup{skip=4pt}
    \captionsetup[subfigure]{justification=centering,skip=2pt}
    \begin{minipage}[t]{0.24\textwidth}
        \centering
        \includegraphics[width=\linewidth]{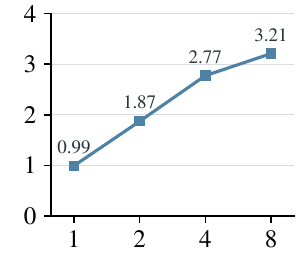}
    \end{minipage}
    \hfill
    \begin{minipage}[t]{0.24\textwidth}
        \centering
        \includegraphics[width=\linewidth]{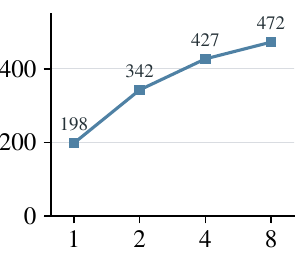}
    \end{minipage}
    \hfill
    \begin{minipage}[t]{0.24\textwidth}
        \centering
        \includegraphics[width=\linewidth]{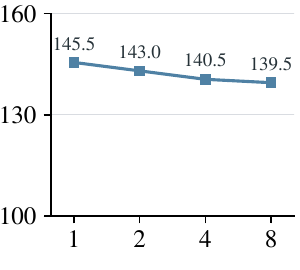}
    \end{minipage}
    \hfill
    \begin{minipage}[t]{0.24\textwidth}
        \centering
        \includegraphics[width=\linewidth]{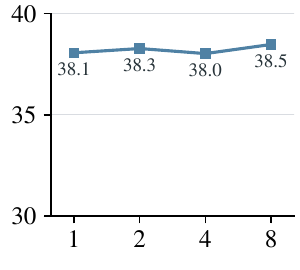}
    \end{minipage}
    \par\vspace{-0.55em}
    {\small Depth}
    \par\vspace{-0.65em}
    \subcaptionbox{Realized depth.}[0.24\textwidth]{\rule{0pt}{0pt}}
    \hfill
    \subcaptionbox{Discarded policy tokens.}[0.24\textwidth]{\rule{0pt}{0pt}}
    \hfill
    \subcaptionbox{Policy decode rate (tok/s).}[0.24\textwidth]{\rule{0pt}{0pt}}
    \hfill
    \subcaptionbox{Agent latency (s).}[0.24\textwidth]{\rule{0pt}{0pt}}
    \caption{Depth sweep with the 8B policy and 0.6B world model. Greater depth increases realized speculation and discarded policy computation without reducing agent latency.}
    \label{fig:depth}
\end{figure*}

Increasing depth allows \method\ to speculate further ahead by repeatedly alternating predicted observations with conditional policy actions. However, in Figure~\ref{fig:depth}, agent latency changes little across $D\in\{1,2,4,8\}$. The pipeline also realizes substantially less depth than configured: increasing $D$ from 1 to 8 raises the mean realized depth from 0.99 to 3.21, since speculation stops whenever tool execution catches up or an incorrect prediction invalidates the speculative work. Deeper speculation also makes rollbacks more expensive because all work downstream of a mismatched observation is discarded. Increasing $D$ from 1 to 8 raises discarded policy tokens from 198 to 472 per rollout. As a result, policy decode rate falls from 145.5 to 139.5 tok/s. Greater depth consequently adds discarded computation and contention without reducing agent latency, making $D=1$ our preferred configuration.

\begin{figure*}[t]
    \centering
    \captionsetup{skip=4pt}
    \captionsetup[subfigure]{justification=centering,skip=2pt}
    \begin{minipage}[t]{0.24\textwidth}
        \centering
        \includegraphics[width=\linewidth]{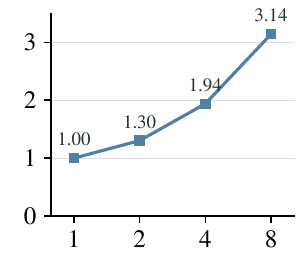}
    \end{minipage}
    \hfill
    \begin{minipage}[t]{0.24\textwidth}
        \centering
        \includegraphics[width=\linewidth]{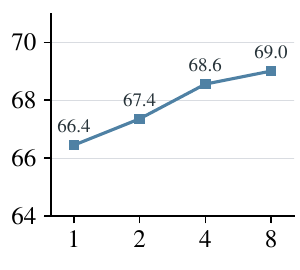}
    \end{minipage}
    \hfill
    \begin{minipage}[t]{0.24\textwidth}
        \centering
        \includegraphics[width=\linewidth]{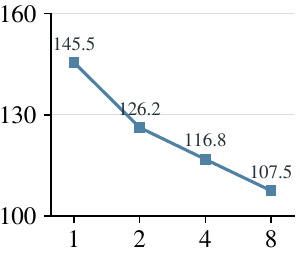}
    \end{minipage}
    \hfill
    \begin{minipage}[t]{0.24\textwidth}
        \centering
        \includegraphics[width=\linewidth]{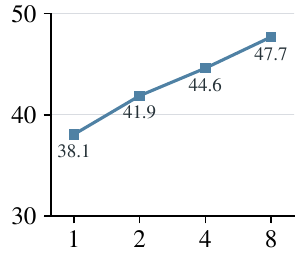}
    \end{minipage}
    \par\vspace{-0.55em}
    {\small Breadth}
    \par\vspace{-0.65em}
    \subcaptionbox{Realized breadth.}[0.24\textwidth]{\rule{0pt}{0pt}}
    \hfill
    \subcaptionbox{Match (\%).}[0.24\textwidth]{\rule{0pt}{0pt}}
    \hfill
    \subcaptionbox{Policy decode rate (tok/s).}[0.24\textwidth]{\rule{0pt}{0pt}}
    \hfill
    \subcaptionbox{Agent latency (s).}[0.24\textwidth]{\rule{0pt}{0pt}}
    \caption{Breadth sweep with the 8B policy and 0.6B world model. Greater breadth increases realized speculation and match rate, while reducing the policy decode rate and increasing agent latency.}
    \label{fig:breadth}
\end{figure*}

\paragraph{Speculation breadth.}

Increasing breadth allows \method\ to sample multiple candidate observations and generate a conditional action for each distinct candidate. However, the additional samples are frequently duplicates. At $K=8$, deduplication leaves only 3.14 distinct candidates per attempt on average, and the share of attempts with a match rises by only 2.6 percentage points over $K=1$. This marginal gain introduces substantial contention, with agent latency increasing by 25.3\% at $K=8$, relative to $K=1$. Over the same range, the policy decode rate falls by 26.1\%, and discarded policy tokens increase by roughly 5x. Thus, $K=1$ is preferred.

\subsection{World Models Should Track the Policy}
\label{sec:staleness-results}

\begin{figure*}[t]
    \centering
    \makebox[\textwidth][r]{\includegraphics[width=0.62\textwidth]{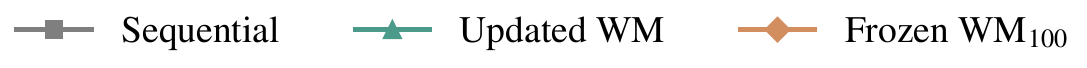}\hspace{0.01\textwidth}}\\[-1.3em]
    \captionsetup{skip=4pt}
    \captionsetup[subfigure]{justification=centering,skip=2pt}
    \makebox[\textwidth][c]{\ignorespaces
    \begin{minipage}[t]{0.32\textwidth}
        \centering
        \includegraphics[width=\linewidth]{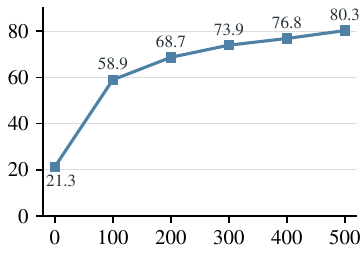}
    \end{minipage}
    \hspace{0.02\textwidth}
    \begin{minipage}[t]{0.32\textwidth}
        \centering
        \includegraphics[width=\linewidth]{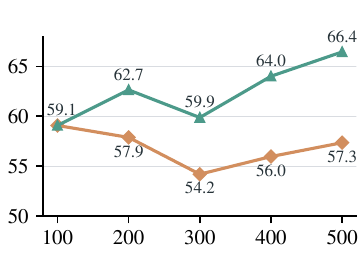}
    \end{minipage}
    \hspace{0.02\textwidth}
    \begin{minipage}[t]{0.32\textwidth}
        \centering
        \includegraphics[width=\linewidth]{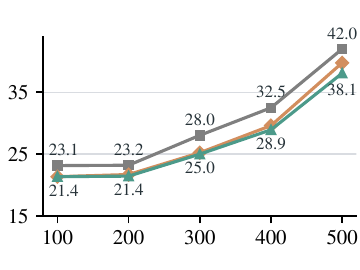}
    \end{minipage}
    \unskip}
    \par\vspace{-0.80em}
    {\small Policy training step}
    \par\vspace{-0.65em}
    \makebox[\textwidth][c]{\ignorespaces
    \subcaptionbox{Reward (\%).\label{fig:staleness-reward}}[0.32\textwidth]{\rule{0pt}{0pt}}
    \hspace{0.02\textwidth}
    \subcaptionbox{Match (\%).\label{fig:staleness-match}}[0.32\textwidth]{\rule{0pt}{0pt}}
    \hspace{0.02\textwidth}
    \subcaptionbox{Agent latency (s).\label{fig:staleness-latency}}[0.32\textwidth]{\rule{0pt}{0pt}}
    \unskip}
    \caption{Policy improvement and the effects of co-training the world model on match rate and agent latency.}
    \label{fig:staleness}
\end{figure*}

As the policy improves (Figure~\ref{fig:staleness-reward}), it changes which tool calls it makes and therefore which tool outputs the world model needs to predict. Co-training the world model with the policy allows the world model to adapt to these changes. To evaluate the benefit of co-training, we compare a world model trained through the same step as the policy with a world model frozen after policy step 100. In Figure~\ref{fig:staleness-match}, the gap in prediction accuracy between the frozen and updated world models widens with the number of policy training steps. The gap grows from 4.8 percentage points at step 200 to 9.1 points at step 500.

This materially affects the agent latency of \method, as seen in Figure~\ref{fig:staleness-latency}. While speculating with either world model consistently outperforms the sequential baseline, the difference in agent latency between the two world models widens during training. Concretely, the latency advantage of the updated world model over the frozen world model grows from 0.3 seconds at policy step 200 to 1.7 seconds at step 500. This widening gap suggests that keeping the world model up to date increases the match rate and decreases agent latency.

\begin{figure*}[t]
    \centering
    \captionsetup{skip=4pt}
    \captionsetup[subfigure]{justification=centering,skip=2pt}
    \makebox[\textwidth][c]{\ignorespaces
    \begin{minipage}[t]{0.32\textwidth}
        \centering
        \includegraphics[width=\linewidth]{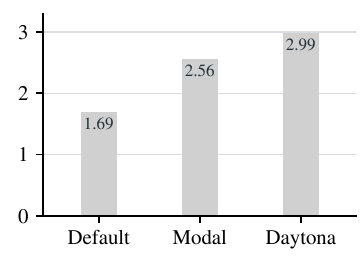}
    \end{minipage}
    \hspace{0.02\textwidth}
    \begin{minipage}[t]{0.32\textwidth}
        \centering
        \includegraphics[width=\linewidth]{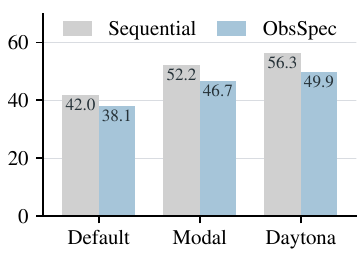}
    \end{minipage}
    \unskip}
    \par\vspace{-1.5em}
    \makebox[\textwidth][c]{\ignorespaces
    \subcaptionbox{Tool latency (s).}[0.32\textwidth]{\rule{0pt}{0pt}}
    \hspace{0.02\textwidth}
    \subcaptionbox{Agent latency (s).}[0.32\textwidth]{\rule{0pt}{0pt}}
    \unskip}
    \caption{Sensitivity to sandbox infrastructure with the 8B policy and 0.6B world model.}
    \label{fig:sandbox}
\end{figure*}

\subsection{\method{} Is Robust to Sandbox Latency}
\label{sec:sandbox-results}

Figure~\ref{fig:sandbox} tests whether \method\ remains effective across sandbox providers with different execution latencies. On the 8B policy, tool latency increases from 1.69 seconds with our default sandbox service to 2.56 seconds on Modal and 2.99 seconds on Daytona. \method\ consistently reduces agent latency, by 9.3\% with our default sandbox service, 10.6\% on Modal, and 11.3\% on Daytona. The gains increase only modestly because speculation at $D=1$ can hide at most one conditional policy generation. With our default sandbox service, the next action finishes generating before the tool returns in 64.6\% of speculation attempts, leaving little additional policy work to overlap during longer tool executions.

\subsection{Not All Tasks Benefit from \method{}}
\label{sec:negative-results}

We also evaluate an additional mathematical reasoning workload: tool-integrated reasoning (TIR)~\citep{gou2024tora,yang2024qwenmath} on competition-level mathematical problems from DAPO-Math-17k~\citep{yu2026dapo}. On this task, the policy interleaves natural-language reasoning with code execution for arithmetic, case enumeration, and intermediate verification before writing its final answer to \texttt{/home/user/answer.txt}. We follow the same experimental configuration as in our main experiments. Agent latency is 55.82 seconds with the sequential pipeline and 56.42 seconds with \method\ using a 0.6B world model.

We identify three reasons why \method\ is ineffective on TIR. First, the workload provides fewer opportunities to speculate than the terminal-use tasks, averaging only 3.44 tool calls per rollout compared with 11.73. Second, tool execution accounts for only 21.2\% of TIR agent latency, versus 47.1\% on the terminal-use tasks, leaving a much smaller fraction of the critical path available for overlap. Third, its observations are harder to predict. With a 0.6B world model, the match rate falls from 66.4\% on the terminal-use tasks to 55.0\% on TIR. Scaling the world model to 8B raises the match rate to 59.5\% but triples prediction latency, leaving agent latency effectively unchanged at 55.79 seconds.

\section{Related Work}
\paragraph{Speculative execution for agents.}
Prior work accelerates tool-using agents by speculating either future actions or future observations. Action speculation predicts and executes future tool calls while the policy generates the current action~\citep{sui2026paste,nichols2025optimizing,bai2026spork,huang2025reducing,zhong2026dualspec,liu2026speculative}. These methods require special care for tools that mutate the environment's state. Observation speculation instead predicts the output of the current tool call and begins the next policy generation before execution finishes. Speculative Actions~\citep{ye2025speculative} increases speculative breadth by predicting multiple candidate outputs from a pending tool call and pre-executing the next call implied by each. SpecHop increases speculative depth across multiple retrieval steps and verifies them asynchronously~\citep{saberi2026spechop}. Speculate with Memory stores memories of prior transitions to improve speculation accuracy~\citep{li2026speculate}. More recently, AOSpec combines action and observation speculation and, like \method, requires the predicted observation to exactly match the actual observation~\citep{chen2026aospec}. \method\ studies observation speculation for RL under three requirements: lossless rollout collection, faster rollout collection under shared policy resources, and accurate predictions as the policy learns. We retain speculative generations only when predicted and actual observations match exactly, limit speculation to manage contention among concurrent rollouts, and co-train the world model with the policy. Our experiments measure how these choices affect rollout latency and throughput.

\paragraph{Efficient RL rollouts.}
An extensive body of work accelerates RL rollouts. Asynchronous systems overlap rollout collection with policy updates~\citep{fu2026areal,piche2025pipelinerl}, and sample-efficient algorithms reduce the number of rollouts required for learning~\citep{chen2025minimax,gao2025soft,shrivastava2026echo}. Other methods accelerate policy inference through low-precision arithmetic~\citep{xiao2023smoothquant,qiu2026fp8}, speculative decoding~\citep{leviathan2023fast,chen2026respec,zhang2026fastgrpo,he2025history,shao2026beat,chang2026srt,liu2025spec,pham2026specroll}, or multi-token prediction~\citep{gloeckle2024better,wang2026mtp}. These methods reduce the time or compute required for RL and can be combined with \method, which overlaps policy generation with environment execution.

\paragraph{World models for agents.}
Prior work uses world models to support planning and agent training. WMA~\citep{chae2025web} and WebDreamer~\citep{gu2024your} predict the results of candidate web actions to improve action selection. WebEvolver uses a co-evolving world model to generate training data for the policy and to simulate future observations during inference~\citep{fang2025webevolver}, while CoMAP uses predicted feedback to refine actions~\citep{liu2026comap}. GTM~\citep{ren2025gtm}, Qwen-AgentWorld~\citep{zuo2026qwen}, WebWorld~\citep{xiao2026webworld}, WebSynthesis~\citep{gao2025websynthesis}, and World Model RL~\citep{yang2026scaling} use world models to simulate the environment for agent training rather than call real tools. ECHO~\citep{shrivastava2026echo}, PaW~\citep{lu2026policy}, and RWML~\citep{yu2026reinforcement} train the policy itself to predict observations as an auxiliary objective. \method\ uses a separate lightweight world model to reduce agent latency. 

\section{Conclusion}
We presented \method, an observation speculation pipeline for agentic RL that overlaps policy generation with environment execution while keeping rollout collection lossless. On terminal-use tasks with 8B and 32B policies, a small co-trained world model reduces both agent and batch latency and increases asynchronous rollout throughput. Our experiments show that increasing speculation depth or breadth increases contention for shared policy resources and can even increase agent latency. As the policy learns, co-training keeps the world model's predictions accurate and reduces agent latency compared with a frozen world model. Together, these results establish observation speculation as a new direction for improving the efficiency of agentic RL. A promising future direction is to speculate selectively based on a tool call's output predictability, expected execution time, and the current utilization of policy resources.

\raggedbottom
\bibliography{iclr2027_conference}
\bibliographystyle{iclr2027_conference}

\appendix
\makeatletter
\setlength{\@fptop}{0pt}
\setlength{\@fpsep}{18pt}
\makeatother
\section{\method\ Algorithm}
\label{sec:rollout-algorithm}

\begin{algorithm}[H]
\caption{\method\ rollout collection for $D=K=1$.}
\label{alg:wmspec}
\begin{algorithmic}[1]
\Statex \textbf{Inputs}\quad instruction $x$, policy $\pi_\theta$, world model $q_\phi$, environment $\mathcal{E}$, initial state $s_1$
\State $t \gets 1$, $h_1 \gets (x)$, and sample $a_t \sim \pi_\theta(\cdot\mid h_t)$
\While{$a_t$ is not \texttt{done} and the rollout budget remains}
    \State Start environment execution $(s_{t+1},o_t) \gets \mathcal{E}(s_t,a_t)$ asynchronously
    \State Start prediction $\widehat{o}_t \sim q_\phi(\cdot\mid h_t,a_t)$ asynchronously
    \State Wait until the environment execution or prediction finishes
    \If{the prediction finishes first}
        \State Start sampling $a_{t+1} \sim \pi_\theta(\cdot\mid h_t,a_t,\widehat{o}_t)$ asynchronously
    \Else
        \State Cancel the prediction
    \EndIf
    \State Await the environment execution to obtain $o_t$
    \State $h_{t+1} \gets (h_t,a_t,o_t)$
    \If{the prediction finished first and $\widehat{o}_t=o_t$}
        \State Await the speculative generation of $a_{t+1}$
    \Else
        \State Cancel speculative generation if it is active
        \State Sample $a_{t+1} \sim \pi_\theta(\cdot\mid h_{t+1})$
    \EndIf
    \State $t \gets t+1$
\EndWhile
\State \Return $(h_t,a_t)$
\end{algorithmic}
\end{algorithm}

\section{Training and Serving Configuration}
\label{sec:training-configuration}

\begin{table}[H]
    \centering
    \small
    \setlength{\tabcolsep}{6pt}
    \renewcommand{\arraystretch}{1.12}
    \begin{tabular}{@{}p{0.30\linewidth}p{0.64\linewidth}@{}}
        \toprule
        Parameter & Value \\
        \midrule
        \multicolumn{2}{@{}l}{\textit{Data}} \\
        Batch size & 16 prompts \\
        Group size & 16 rollouts per prompt \\
        Prompt budget & 2,048 tokens \\
        Response budget & 16,384 tokens \\
        Per-turn generation budget & 2,048 tokens \\
        Max turns & 32 turns \\
        \addlinespace
        \multicolumn{2}{@{}l}{\textit{Policy optimization}} \\
        Objective & GRPO \\
        Optimizer & AdamW \\
        Learning rate & $1\times10^{-6}$ \\
        Learning-rate schedule & Constant with no warmup \\
        Adam $\left(\beta_1,\beta_2\right)$ & $(0.9,0.999)$ \\
        Weight decay & $0.01$ \\
        Gradient clipping & $1.0$ \\
        Loss aggregation & token-mean \\
        PPO epochs & 1 \\
        GRPO clip range & $[0.20,0.28]$ \\
        KL coefficient & $0.0$ \\
        Entropy coefficient & $0.0$ \\
        Training steps & 500 \\
        \addlinespace
        \multicolumn{2}{@{}l}{\textit{World model optimization}} \\
        Objective & Masked cross-entropy \\
        Optimizer & AdamW \\
        Learning rate & $1\times10^{-5}$ \\
        Learning-rate schedule & Constant with no warmup \\
        Adam $\left(\beta_1,\beta_2\right)$ & $(0.9,0.999)$ \\
        Weight decay & $0.01$ \\
        Gradient clipping & $1.0$ \\
        Loss aggregation & token-mean \\
        Training steps & 500 \\
        \addlinespace
        \multicolumn{2}{@{}l}{\textit{Policy serving}} \\
        Serving engine & vLLM \\
        TP size & 2 \\
        Temperature & $1.0$ \\
        Top-$p$ & $1.0$ \\
        \addlinespace
        \multicolumn{2}{@{}l}{\textit{World model serving}} \\
        Serving engine & vLLM \\
        TP size & 1 \\
        Temperature & $0.0$ (or $1.0$ for additional candidates when $K>1$) \\
        Top-$p$ & $1.0$ \\
        \bottomrule
    \end{tabular}
    \caption{Training and serving configuration for the policy and world model.}
    \label{tab:training-configuration}
\end{table}

\clearpage
\section{Full Results}
\label{sec:full-results}

\begin{table*}[!ht]
    \centering
    \small
    \setlength{\tabcolsep}{4pt}
    \begin{tabular}{llcccc}
        \toprule
        Policy & Pipeline & Reward (\%) $\uparrow$ & Agent (s) $\downarrow$ & Batch (s) $\downarrow$ & Rollouts/h $\uparrow$ \\
        \midrule
        \multirow{2}{*}{Qwen3-8B}
            & Sequential & $78.64 \pm 0.64$ & $41.98 \pm 0.39$ & $156.26 \pm 2.94$ & $13{,}762 \pm 156$ \\
            & \method    & $79.76 \pm 0.63$ & $38.06 \pm 0.38$ & $149.42 \pm 3.06$ & $14{,}571 \pm 175$ \\
        \midrule
        \multirow{2}{*}{Qwen3-32B}
            & Sequential & $84.62 \pm 0.56$ & $42.59 \pm 0.46$ & $185.63 \pm 2.40$ & $14{,}066 \pm 155$ \\
            & \method    & $83.91 \pm 0.57$ & $36.53 \pm 0.44$ & $171.16 \pm 2.81$ & $15{,}689 \pm 199$ \\
        \bottomrule
    \end{tabular}
    \caption{Full results underlying Figure~\ref{fig:main-results}.}
    \label{tab:main-results}
\end{table*}

\begin{table*}[!ht]
    \centering
    \small
    \setlength{\tabcolsep}{4pt}
    \resizebox{\ifdim\width>\linewidth\linewidth\else\width\fi}{!}{
    \begin{tabular}{lcccccc}
        \toprule
        Command & Count & Tool (s) & WM (s) & Match (\%) & Rollback (\%) & Cancel (\%) \\
        \midrule
        \texttt{cat} & 19,544 & $1.597 \pm 0.006$ & $0.113 \pm 0.001$ & $79.95 \pm 0.29$ & $16.75 \pm 0.27$ & $3.31 \pm 0.13$ \\
        \texttt{ls} & 10,681 & $1.614 \pm 0.008$ & $0.121 \pm 0.001$ & $39.43 \pm 0.47$ & $59.54 \pm 0.47$ & $1.03 \pm 0.10$ \\
        \texttt{mkdir} & 2,943 & $1.607 \pm 0.016$ & $0.036 \pm 0.000$ & $96.67 \pm 0.33$ & $3.33 \pm 0.33$ & $0.00 \pm 0.00$ \\
        \texttt{echo} & 1,169 & $1.612 \pm 0.026$ & $0.038 \pm 0.000$ & $98.37 \pm 0.37$ & $1.63 \pm 0.37$ & $0.00 \pm 0.00$ \\
        \texttt{sha256sum} & 946 & $1.625 \pm 0.028$ & $0.178 \pm 0.002$ & $35.20 \pm 1.55$ & $50.00 \pm 1.63$ & $14.80 \pm 1.15$ \\
        \texttt{sqlite3} & 668 & $1.631 \pm 0.033$ & $0.102 \pm 0.004$ & $78.89 \pm 1.58$ & $15.57 \pm 1.40$ & $5.54 \pm 0.89$ \\
        \texttt{make} & 664 & $1.766 \pm 0.038$ & $0.170 \pm 0.016$ & $42.02 \pm 1.92$ & $47.74 \pm 1.94$ & $10.24 \pm 1.18$ \\
        \texttt{date} & 412 & $1.541 \pm 0.044$ & $0.063 \pm 0.001$ & $0.24 \pm 0.24$ & $99.76 \pm 0.24$ & $0.00 \pm 0.00$ \\
        \texttt{find} & 357 & $1.501 \pm 0.046$ & $0.081 \pm 0.004$ & $38.94 \pm 2.58$ & $30.81 \pm 2.44$ & $30.25 \pm 2.43$ \\
        \texttt{python3} & 338 & $8.011 \pm 0.737$ & $0.200 \pm 0.010$ & $48.52 \pm 2.72$ & $45.56 \pm 2.71$ & $5.92 \pm 1.28$ \\
        \midrule
        Overall & 47,762 & $1.685 \pm 0.008$ & $0.105 \pm 0.001$ & $66.45 \pm 0.22$ & $30.50 \pm 0.21$ & $3.05 \pm 0.08$ \\
        \bottomrule
    \end{tabular}
    }
    \caption{Full results underlying Figure~\ref{fig:command-predictability}.}
    \label{tab:command-predictability}
\end{table*}

\begin{table*}[!ht]
    \centering
    \small
    \setlength{\tabcolsep}{4pt}
    \begin{tabular}{lccccc}
        \toprule
        WM & WM latency (s) & Agent latency (s) & Match & Rollback & Cancel \\
        \midrule
        0.6B & $0.105 \pm 0.001$ & $38.06 \pm 0.38$ & $66.45 \pm 0.22$\% & $30.50 \pm 0.21$\% & $3.05 \pm 0.08$\% \\
        1.7B & $0.137 \pm 0.001$ & $37.84 \pm 0.37$ & $68.50 \pm 0.21$\% & $28.71 \pm 0.21$\% & $2.80 \pm 0.08$\% \\
        4B   & $0.239 \pm 0.001$ & $37.89 \pm 0.36$ & $68.74 \pm 0.21$\% & $24.88 \pm 0.20$\% & $6.38 \pm 0.11$\% \\
        8B   & $0.350 \pm 0.002$ & $37.95 \pm 0.37$ & $66.17 \pm 0.22$\% & $22.47 \pm 0.19$\% & $11.36 \pm 0.15$\% \\
        \bottomrule
    \end{tabular}
    \caption{Full results underlying Figure~\ref{fig:wm-size}.}
    \label{tab:wm-size}
\end{table*}

\begin{table*}[!ht]
    \centering
    \small
    \setlength{\tabcolsep}{3pt}
    \begin{tabular}{lcccc}
        \toprule
        Depth & Agent latency (s) & Realized depth & Policy tok/s & Discarded policy tok. \\
        \midrule
        1 & $38.06 \pm 0.38$ & $0.992 \pm 0.001$ & $145.5 \pm 0.1$ & $198.0 \pm 3.4$ \\
        2 & $38.27 \pm 0.38$ & $1.871 \pm 0.006$ & $143.0 \pm 0.1$ & $342.5 \pm 5.8$ \\
        4 & $38.02 \pm 0.37$ & $2.773 \pm 0.017$ & $140.5 \pm 0.1$ & $426.7 \pm 7.5$ \\
        8 & $38.47 \pm 0.39$ & $3.208 \pm 0.029$ & $139.5 \pm 0.1$ & $472.2 \pm 9.8$ \\
        \bottomrule
    \end{tabular}
    \caption{Full results underlying Figure~\ref{fig:depth}.}
    \label{tab:depth}
\end{table*}

\begin{table*}[!ht]
    \centering
    \small
    \setlength{\tabcolsep}{3pt}
    \resizebox{\ifdim\width>\linewidth\linewidth\else\width\fi}{!}{
    \begin{tabular}{lccccc}
        \toprule
        $K$ & Agent latency (s) & Match & Realized breadth & Policy tok/s & Discarded policy tok. \\
        \midrule
        1 & $38.06 \pm 0.38$ & $66.45 \pm 0.22$\% & 1.0 & $145.5 \pm 0.1$ & $198.0 \pm 3.4$ \\
        2 & $41.88 \pm 0.40$ & $67.36 \pm 0.22$\% & $1.303 \pm 0.002$ & $126.2 \pm 0.1$ & $332.8 \pm 5.8$ \\
        4 & $44.60 \pm 0.44$ & $68.56 \pm 0.21$\% & $1.940 \pm 0.006$ & $116.8 \pm 0.1$ & $574.9 \pm 10.3$ \\
        8 & $47.68 \pm 0.46$ & $69.01 \pm 0.21$\% & $3.142 \pm 0.013$ & $107.5 \pm 0.1$ & $994.9 \pm 19.0$ \\
        \bottomrule
    \end{tabular}
    }
    \caption{Full results underlying Figure~\ref{fig:breadth}.}
    \label{tab:breadth}
\end{table*}

\begin{table*}[!ht]
    \centering
    \small
    \setlength{\tabcolsep}{3pt}
    \resizebox{\ifdim\width>\linewidth\linewidth\else\width\fi}{!}{
    \begin{tabular}{ccccccc}
        \toprule
        Policy step & Reward (\%) & Updated match (\%) & Frozen match (\%) & Sequential agent (s) & Updated agent (s) & Frozen agent (s) \\
        \midrule
        100 & $58.95 \pm 2.52$ & $59.07 \pm 0.23$ & $59.07 \pm 0.23$ & $23.14 \pm 0.42$ & $21.35 \pm 0.40$ & $21.35 \pm 0.40$ \\
        200 & $68.68 \pm 2.38$ & $62.64 \pm 0.23$ & $57.88 \pm 0.24$ & $23.20 \pm 0.27$ & $21.44 \pm 0.26$ & $21.70 \pm 0.30$ \\
        300 & $73.95 \pm 2.25$ & $59.87 \pm 0.22$ & $54.19 \pm 0.22$ & $28.03 \pm 0.33$ & $24.99 \pm 0.30$ & $25.22 \pm 0.28$ \\
        400 & $76.84 \pm 2.16$ & $64.01 \pm 0.21$ & $55.96 \pm 0.21$ & $32.47 \pm 0.31$ & $28.92 \pm 0.28$ & $29.60 \pm 0.29$ \\
        500 & $80.26 \pm 2.04$ & $66.45 \pm 0.22$ & $57.35 \pm 0.23$ & $41.98 \pm 0.39$ & $38.06 \pm 0.38$ & $39.74 \pm 0.39$ \\
        \bottomrule
    \end{tabular}
    }
    \caption{Full results underlying Figure~\ref{fig:staleness}.}
    \label{tab:staleness}
\end{table*}

\begin{table*}[!ht]
    \centering
    \small
    \setlength{\tabcolsep}{4pt}
    \begin{tabular}{lcccc}
        \toprule
        Sandbox & Tool latency (s) & Sequential agent (s) & \method\ agent (s) & Reduction \\
        \midrule
        Default  & $1.69 \pm 0.01$ & $41.98 \pm 0.39$ & $38.06 \pm 0.38$ & 9.3\% \\
        Modal    & $2.56 \pm 0.01$ & $52.17 \pm 0.44$ & $46.66 \pm 0.51$ & 10.6\% \\
        Daytona  & $2.99 \pm 0.02$ & $56.30 \pm 0.59$ & $49.94 \pm 0.54$ & 11.3\% \\
        \bottomrule
    \end{tabular}
    \caption{Full results underlying Figure~\ref{fig:sandbox}. All measured quantities report means and standard errors.}
    \label{tab:sandbox}
\end{table*}

\begin{table*}[!ht]
    \centering
    \small
    \setlength{\tabcolsep}{4pt}
    \resizebox{\ifdim\width>\linewidth\linewidth\else\width\fi}{!}{
    \begin{tabular}{lcccccc}
        \toprule
        Method & Reward (\%) & Agent (s) & WM (s) & Match (\%) & Rollback (\%) & Cancel (\%) \\
        \midrule
        Sequential & $78.25 \pm 0.64$ & $55.82 \pm 0.83$ & -- & -- & -- & -- \\
        \method, 0.6B & $78.47 \pm 0.64$ & $56.42 \pm 0.86$ & $0.120 \pm 0.003$ & $54.96 \pm 0.42$ & $40.42 \pm 0.41$ & $4.62 \pm 0.18$ \\
        \method, 8B   & $78.39 \pm 0.64$ & $55.79 \pm 0.82$ & $0.367 \pm 0.007$ & $59.47 \pm 0.41$ & $31.72 \pm 0.39$ & $8.81 \pm 0.24$ \\
        \bottomrule
    \end{tabular}
    }
    \caption{Full results for the TIR experiments in Section~\ref{sec:negative-results}.}
    \label{tab:tir}
\end{table*}

\end{document}